\documentclass[nonacm,sigconf]{acmart}
\renewcommand\footnotetextcopyrightpermission[1]{} 
\usepackage{hyperref}
\usepackage[anythingbreaks]{breakurl}

\def\BibTeX{{\rm B\kern-.05em{\sc i\kern-.025em b}\kern-.08emT\kern-.1667em\lower.7ex\hbox{E}\kern-.125emX}}
    
\setcopyright{none}

\usepackage[utf8]{inputenc}    
\usepackage[T1]{fontenc}        
\usepackage{hyperref}           
\usepackage{url}                    
\usepackage{booktabs}           
\usepackage{amsfonts}           
\usepackage{nicefrac}           
\usepackage{microtype}          
\usepackage{longtable}
\usepackage{graphicx}
\usepackage{subcaption}
\usepackage{algorithm}
\usepackage{capt-of}
\usepackage{xcolor}
\usepackage{mathtools}        
\usepackage{algpseudocode}
\usepackage{enumitem}

\usepackage{pgfplots}
\usepackage{tikz}
\usetikzlibrary{calc,fadings,shapes.misc,3d,arrows,
decorations,decorations.pathmorphing,
decorations.text,shapes.geometric,intersections,
decorations.markings,patterns,spy,chains,positioning}
\usepgfplotslibrary{fillbetween,polar,patchplots}

\begin{document}

\title{Semi-Supervised Learning for Bilingual Lexicon Induction}

%
\author{Paul Garnier}
\affiliation{%
  \institution{MINES Paristech , PSL - Research University}
  \country{France}
  \authornote{paul.garnier@minesparis.psl.eu}
}
\author{Gauthier Guinet}

\author{May 2020}



\begin{abstract}
\textbf{
We consider the problem of aligning two sets of continuous word representations, corresponding to languages, to a common space in order to infer a bilingual lexicon. It was recently shown that it is possible to infer such lexicon, without using any parallel data, by aligning word embeddings trained on monolingual data. Such line of work is called unsupervised bilingual induction.  By wondering whether it was possible to gain experience in the progressive learning of several languages, we asked ourselves to what extent we could integrate the knowledge of a given set of languages when learning a new one, without having parallel data for the latter. In other words, while keeping the core problem of unsupervised learning in the latest step, we allowed the access to other corpora of idioms, hence the name semi-supervised. This led us to propose a novel formulation, considering the lexicon induction as a ranking problem for which we used recent tools of this machine learning field. Our experiments on standard benchmarks, inferring dictionary from English to more than 20 languages, show that our approach consistently outperforms existing state of the art benchmark. In addition, we deduce from this new scenario several relevant conclusions allowing a better understanding of the alignment phenomenon.  
}
\\
\end{abstract}

\keywords{Bilingual Unsupervised Alignment, Word Embeddings, Learning to Rank, Wasserstein Procustes}
\maketitle
\section{Introduction}

Word vectors are conceived to synthesize and quantify semantic nuances, using a few hundred coordinates. These are generally used in downstream tasks to improve generalization when the amount of data is scarce. The widespread use and successes of these "word embeddings" in monolingual tasks has inspired further research on the induction of multilingual word embeddings for two or more languages in the same vector space. 

The starting point was the discovery  \citet{bib26} that word embedding spaces have similar structures across languages, even when considering distant language pairs like English and Vietnamese. More precisely, two sets of pre-trained vectors in different languages can be aligned to some extent: good word translations can be produced through a simple linear mapping between the two sets of embeddings. As an example, learning a direct mapping between Italian and Portuguese leads to a word translation accuracy of 78.1\% with a nearest neighbor (NN) criterion.

Embeddings of translations and words with similar meaning are close (geometrically) in the shared  cross-lingual vector space. This property makes them very effective for cross-lingual Natural Language Processing (NLP) tasks.
The simplest way to evaluate the result is the Bilingual Lexicon Induction (BLI) criterion, which designs the percentage of the dictionary that can be correctly induced. 
Thus, BLI is often the first step towards several downstream tasks such as Part-Of-Speech (POS) tagging, parsing, document classification, language genealogy analysis or (unsupervised) machine translation.

Frequently, these common representations are learned through a two-step process, whether in a bilingual or multilingual setting. First, monolingual word representations are learned over large portions of text; these pre-formed representations are actually available for several languages and are widely used, such as the Fasttext Wikipedia vectors used in this work. Second, a correspondence between languages is learned in three ways: in a supervised manner if parallel dictionaries or data are available to be used for supervisory purposes, with minimal supervision, for example by using only identical strings, or in a completely unsupervised manner.\\
\newline
It is common practice in the literature on the subject to separate these two steps and not to address them simultaneously in a paper. Indeed, measuring the efficiency of the algorithm would lose its meaning if the corpus of vectors is not identical at the beginning. We will therefore use open-source data from Facebook containing embeddings of several dozen languages computed using Wikipedia data. The fact that the underlying text corpus is identical also helps to reinforce the isomorphic character of the point clusters. \\
\newline
Concerning the second point, although three different approaches exist, they are broadly based on the same ideas: the goal is to identify a subset of points that are then used as anchors points to achieve alignment. In the supervised approach, these are the words for which the translation is available. In the semi-supervised approach, we will gradually try to enrich the small initial corpus to have more and more anchor points. The non-supervised approach differs because there is no parallel corpus or dictionary between the two languages. The subtlety of the algorithms will be to release a potential dictionary and then to enrich it progressively. \\

We will focus in the following work on this third approach. Although it is a less frequent scenario, it is of great interest for several reasons. First of all, from a theoretical point of view, it provides a practical answer to a very interesting problem of information theory: given a set of texts in a totally unknown language, what information can we retrieve? The algorithms we chose to implement contrast neatly with the classical approach used until now. Finally, for very distinct languages or languages that are no longer used, it is true that the common corpus can be very thin. 

Many developments have therefore taken place in recent years in this field of unsupervised bilingual lexicon induction. One of the recent discoveries, which pushed us to do this project, is the idea that using information from other languages during the training process helps improve translating language pairs. We came across this idea while searching for multi-alignment of languages instead of bi-alignment. 

We chose to approach the problem in a different way, keeping an unsupervised alignment basis however.  We asked ourselves to what extent we could integrate the knowledge of a given set of languages when learning a new one, without having parallel data for the latter. The scenario of multi-alignment unsupervised assumes that there is no parallel data for all language pairs. We think it is more realistic and useful to assume that we do not have this data only for the last language. 

The underlying learning theme led us to formulate the problem as follows:\textbf{ is it possible to gain experience in the progressive learning of several languages?} In other words, how can we make good use of the learning of several acquired languages to learn a new one? To our knowledge, this problem has never yet been addressed in the literature on the subject. This new formulation led us to consider the lexicon induction as a ranking problem for which we used recent tools of this machine learning field called Learning to Rank. 

In summary, this paper make the following main contributions:
\begin{enumerate}
    \item We present a new approach for the unsupervised bilingual lexicon induction problem that consistently outperforms state-of-the-art methods on several language pairs. On a standard word translation retrieval benchmark, using 200k vocabularies, our method reaches 95.3\% accuracy on English-Spanish while the best unsupervised approach is at 84.1\%. By doing this, we set a new benchmark in the field. 
    \item We conduced a study on the impact of the idioms used for the learning and for the prediction step, allowing us to have a better core understanding of our approach and to forecast the efficiency for a new idiom. 
    \item Our results further strengthen in a new way the strong hypothesis that word embedding spaces have similar structures across languages. \citet{bib26}
\end{enumerate}

We will proceed as follows:  Sections ~\ref{sec:state_art} and ~\ref{sec:eval_mtri} will outline  the state of the art and the different techniques used for unsupervised learning in this context. In particular, we will explain the Wasserstein Procustes approach for bilingual and multi alignment. We then emphasize the lexicon induction given the alignment. Section ~\ref{sec:ltr} presents the Learning to Rank key concepts alongside the TensorFlow framework used in our main algorithm. Section ~\ref{sec:semi_sup} describes our program, the different subtleties and the key parameters. Finally, Section ~\ref{sec:results} presents the experimental results we obtained.


\section{Unsupervised Bilingual Alignement} \label{sec:state_art}

In this section, we provide a brief overview of unsupervised bilingual alignment methods to learn a mapping between two sets of embeddings. The majority are divided into two stages: the actual alignment and lexicon induction, given the alignment. Even if the lexicon induction is often taken into account when aligning (directly or indirectly, through the loss function), this distinction is useful from a theoretical point of view. 
\begin{figure}[ht]
\begin{center}
\includegraphics[scale=0.25]{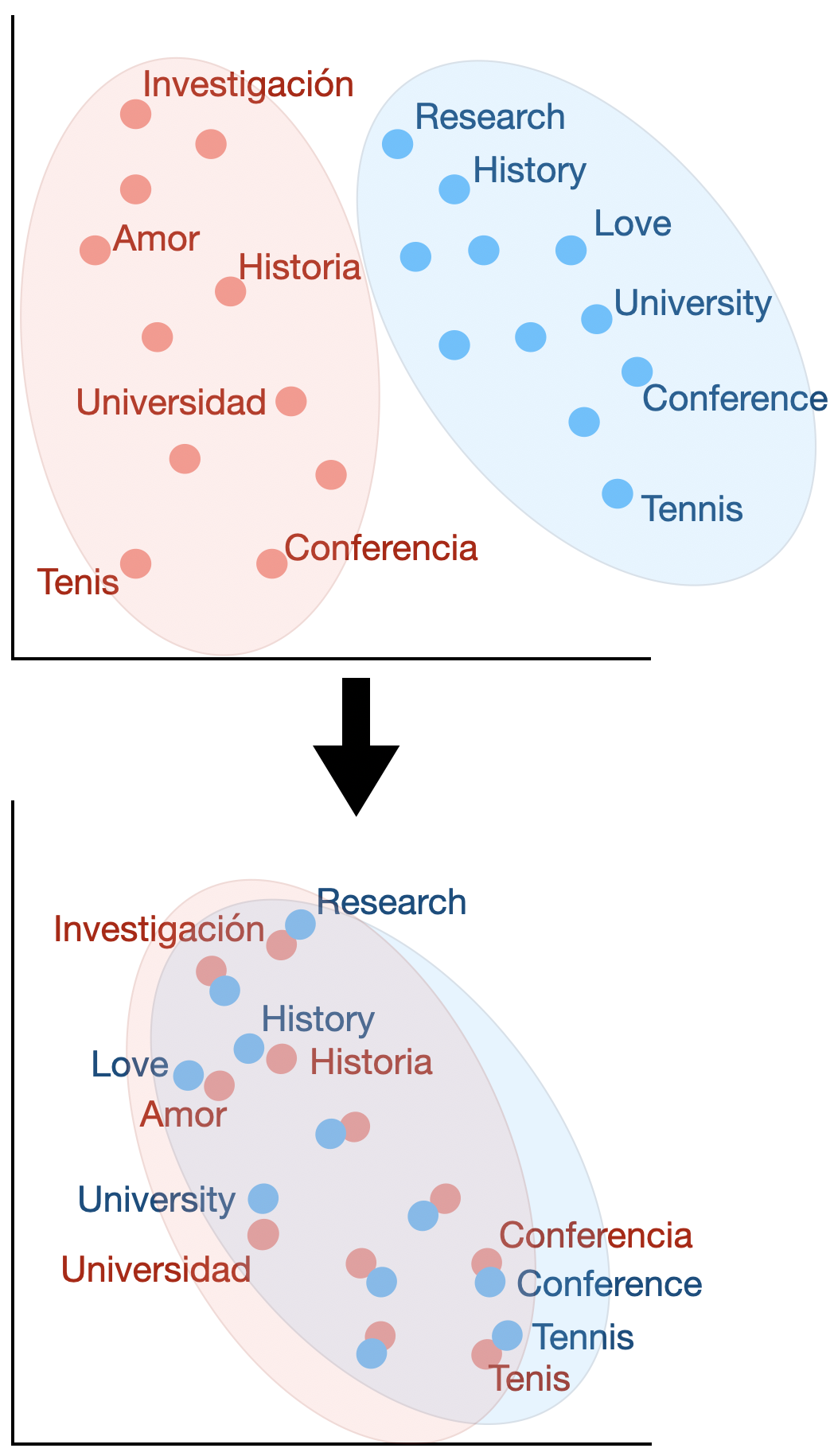}
\end{center}
\caption{Word embeddings alignment (in dimension 2).}
\label{fig:rank}
\end{figure}
Historically, the problem of word vector alignment has been formulated as as a quadratic problem.
This approach, resulting from the supervised literature then allowed to presume the absence of lexicon without modifying to much the framework. That is why we will deal with it first in what follows.

\subsection{Orthogonal Procrustes Problem}

Procustes is a method that aligns points if given the correspondences between them (supervised scenario).
$\mathbf{X} \in \mathbb{R}^{n \times d}$ and $\mathbf{Y} \in \mathbb{R}^{n \times d}$ are the two sets of word embeddings or points and we suppose, as previously said, that we know which point \textbf{X} corresponds to which point $\mathbf{Y}$. This leads us to solve the following least-square problem of optimization, looking for the $\mathbf{W}$ matrix performing the alignment \citet{bib2} :

$$\min _{\mathbf{W} \in \mathbb{R}^{d \times d}}\|\mathbf{X} \mathbf{W}-\mathbf{Y}\|_{2}^{2}$$ 
\newline
We have access to a closed form solution with a cubic complexity. 
Restraining $\mathbf{W}$ to the set of orthogonal matrices $\mathcal{O}_{d}$, improves the alignments for two reasons: it limits overfitting by reducing the size of the minimization space and allows to translate the idea of keeping distances and angles, resulting from the similarity in the space structure. The resulting problem is known as Orthogonal Procrustes and it also admits a closed form solution through a singular value decomposition (cubic complexity).

Thus, if their correspondences are known, the translation matrix between two sets of points can be inferred without too much difficulties. The next step leading to unsupervised learning is to discover these point correspondences using Wasserstein distance.

\subsection{Wasserstein Distance}

In a similar fashion, finding the correct mapping between two sets of word can be done by solving the following minimization problem: 

$$\min _{\mathbf{P} \in \mathcal{P}_{n}}\|\mathbf{X}-\mathbf{P} \mathbf{Y}\|_{2}^{2}$$

$\mathcal{P}_{n}$ containing all the permutation matrices, the solution of the minimization, {Pt} will be an alignment matrix giving away the pair of words. This 1 to 1 mapping can be achieved thanks to the Hungarian algorithm.
It is equivalent to solve the following linear program: $$\max _{\mathbf{P} \in \mathcal{P}_{n}} \operatorname{tr}\left(\mathbf{X}^{\top} \mathbf{P} \mathbf{Y}\right)$$

The combination of the Procustes- Wasserstein minimization problem is the following: $$\min _{\mathbf{Q} \in \mathcal{O}_{d}} \min _{\mathbf{P} \in \mathcal{P}_{n}}\|\mathbf{X Q}-\mathbf{P} \mathbf{Y}\|_{2}^{2}$$

In order to solve this problem, the approach of \citet{bib2} was to use a stochastic optimization algorithm. 
As solving separately those 2 problems was leading to bad local optima, their choice was to select a smaller batch of size \textit{b}, and perform their minimization algorithm on these sub-samples. The batch is playing the role of anchors points. 

\subsection{Multilingual alignment} 

A natural way to improve the efficiency of these algorithms is to consider more than 2 languages.
Thus,when it comes to aligning multiple languages together, two principle approaches quickly come to mind and correspond to two types of optimization problems:

\begin{enumerate}
    \item Align all languages to one pivot language, often English, without taking into account for the loss function other alignments. This leads to low complexity but also to low efficiency between the very distinct language, forced to transit through English. 
    \item Align all language pairs, by putting them all in the loss function, without giving importance to any one in particular. If this improves the efficiency of the algorithm, the counterpart is in the complexity, which is very important because it is quadratic in the number of languages.
\end{enumerate}

A trade-off must therefore be found between these two approaches.

Let us consider $\mathbf{X}_{i}$ word embeddings for each language \textit{i}, \textit{i}=0 can be considered as the reference language, $\mathbf{W}_{i}$ is the mapping matrix we want to learn and $\mathbf{P}_{i}$ the permutation matrix. The alignment of multiple languages using a reference language as pivot can be resumed by the following problem:
$$\min _{\mathbf{W}_{i} \in \mathcal{O}_{d}, \mathbf{P}_{i} \in \mathcal{P}_{n}} \sum_{i} \ell\left(\mathbf{X}_{i} \mathbf{W}_{i}, \mathbf{P}_{i} \mathbf{X}_{0}\right)$$
As said above, although this method gives satisfying results concerning the translations towards the reference language, it provides poor alignment for the secondary languages between themselves.
\newline
Therefore an interesting way of jointly aligning multiple languages to a common space has been brought through by Alaux and al.\citet{bib1}.
\newline
The idea is to consider each interaction between two given languages, therefore the previous sum becomes a double sum with two indexes \textit{i} and \textit{j}. To prevent the complexity from being to high and in order to keep track and control over the different translations, each translation between two languages is given a weight $\alpha_{i j}$ :
$$\min _{\mathbf{Q}_{i} \in \mathcal{O}_{d}, \mathbf{P}_{i j} \in \mathcal{P}_{n}} \sum_{i, j} \alpha_{i j} \ell\left(\mathbf{X}_{i} \mathbf{Q}_{i}, \mathbf{P}_{i j} \mathbf{X}_{j} \mathbf{Q}_{j}\right)$$
The choice of these weights depends on the importance we want to give to the translation from language \textit{i} to language \textit{j}. 
\newline
The previous knowledge we have on the similarities between two languages can come at hand here, for they will have a direct influence on the choice of the weight. However choosing the appropriate weights can be uneasy. For instance giving a high weight to a pair of close languages can be unnecessary and doing the same for two distant languages can be a waste of computation. In order to reach this minimization effectively, they use an algorithm very similar to the stochastic optimization algorithm described above. 

At the beginning, we wanted to use this algorithm to incorporate exogenous knowledge about languages to propose constants $\alpha_{i j}$ more relevant and leading to greater efficiency. Different techniques could result in these parameters: from the mathematical literature such as the Gromov-Wasserstein distance evoked above or from the linguistic literature, using the etymological tree of languages to approximate their degree of proximity or even from both. In the article implementing this algorithm, it is however specified that the final $\alpha_{i j}$ are actually very simple: they are N if we consider a link to the pivot or 1 otherwise. Practical simulations have also led us to doubt the efficiency of this idea. This is why we decided to focus on the idea below that seemed more promising rather than on multialignments. 

\section{Word translation as a retrieval task: Post-alignment Lexicon Induction} \label{sec:eval_mtri}

The core idea of the least-square problem of optimization in Wasserstein Procustes is to minimize the distance between a word and its translation. Hence, given the alignment, the inference part first just consisted in finding the nearest neighbors (NN). Yet, this criterion had a mayor issue:
Nearest neighbors are by nature asymmetric: y being a K-NN of x does not imply that x is a K-NN of y. In high-dimensional spaces, this leads to a phenomenon that is detrimental to matching pairs based on a nearest neighbor rule: some vectors, called hubs, are with high probability nearest neighbors of many other points, while others (anti-hubs) are not nearest neighbors of any point. \citet{bib28}
\newline
Two solutions to this problem have been brought through new criteria, aiming at giving similarity measure between two embeddings, thus allowing to match them appropriately. Among them, the most popular is Cross-Domain Similarity Local Scaling (CSLS) \citet{bib5}. Other exist such as Inverted Softmax (ISF)\citet{bib4}, yet they usually require to estimate noisy parameter in an unsupervised setting where we do not have a direct cross-validation criterion. 

The idea behind CSLS is quite simple: it is a matter of calculating a cosine similarity between the two vectors, subtracting a penalty if one or both of the vectors is also similar at many other points. More formally, we denote by \(\mathcal{N}_{\mathrm{T}}\left({W} x_{s}\right)\) the neighboors of \(\boldsymbol{x}_{\boldsymbol{S}}\) for the target language, after the alignment (hence the presence of 
$\mathbf{W})$.  Similarly we denote by \(\mathcal{N}_{\mathrm{S}}\left(y_{t}\right)\) the neighborhood associated with a word \(t\) of the target language. The penalty term we consider is the mean similarity of a source embedding \(x_{s}\) to its target neighborhood:
$$
r_{\mathrm{T}}\left(W x_{s}\right)=\frac{1}{K} \sum_{y_{t} \in \mathcal{N}_{\mathrm{T}}\left(W x_{s}\right)} \cos \left(W x_{s}, y_{t}\right)
$$
where cos(...) is the cosine similarity. Likewise we denote by \(r_{\mathrm{S}}\left(y_{t}\right)\) the mean similarity of a target word \(y_{t}\) to its neighborhood. Finally, the CSLS is defined as:
$$
\operatorname{CSLS}\left(W x_{s}, y_{t}\right)=2 \cos \left(W x_{s}, y_{t}\right)-r_{\mathrm{T}}\left(W x_{s}\right)-r_{\mathrm{S}}\left(y_{t}\right)
$$

However, it may seem irrelevant to align the embedding words with the NN criterion metric and to use the CSLS criterion in the inference phase. Indeed, it creates a discrepancy between the learning of the translation model and the inference: the global minimum on the set of vectors of one does not necessarily correspond to the one of the other. This naturally led to modify the least-square optimization problem to propose a loss function associated with CSLS \citet{bib27}. By assuming that word vectors are \(\ell_{2}-\) normalized, we have \(\cos \left(\mathbf{W} \mathbf{x}_{i}, \mathbf{y}_{i}\right)=\mathbf{x}_{i}^{\top} \mathbf{W}^{\top} \mathbf{y}_{i} .\) Similarly, we have
$$
\left\|\mathbf{y}_{j}-\mathbf{W} \mathbf{x}_{i}\right\|_{2}^{2}=2-2 \mathbf{x}_{i}^{\top} \mathbf{W}^{\top} \mathbf{y}_{j} .
$$
Therefore, finding the \(k\) nearest neighbors of \(\mathbf{W} \mathbf{x}_{i}\) among the elements of \(\mathbf{Y}\) is equivalent to finding the \(k\) elements of \(\mathbf{Y}\) which have the largest dot product with \(\mathbf{W} \mathbf{x}_{i}\). This equivalent formulation is adopted because it leads to a convex formulation when relaxing the orthogonality constraint on \(\mathbf{W}\). This optimization problem with the Relaxed CSLS loss (RCSLS) is written as:
$$
\begin{array}{c}
\min _{\mathbf{W} \in \mathcal{O}_{d}} \frac{1}{n} \sum_{i=1}^{n}-2 \mathbf{x}_{i}^{\top} \mathbf{W}^{\top} \mathbf{y}_{i} \\
\quad+\frac{1}{k} \sum_{\mathbf{y}_{j} \in \mathcal{N}_{Y}\left(\mathbf{W} \mathbf{x}_{i}\right)} \mathbf{x}_{i}^{\top} \mathbf{W}^{\top} \mathbf{y}_{j} \\
+\frac{1}{k} \sum_{\mathbf{W} \mathbf{x}_{j} \in \mathcal{N}_{X}\left(\mathbf{y}_{i}\right)} \mathbf{x}_{j}^{\top} \mathbf{W}^{\top} \mathbf{y}_{i}
\end{array}
$$
A convex relaxation can then be computed, by considering the convex hull of $\mathcal{O}_{d}$, i.e., the unit ball of the spectral norm. The results of the papers \citet{bib5} point out that RCSLS outperforms the state of the art by, on average, 3 to 4\% in accuracy compared to benchmark. This shows the importance of using the same criterion during training and inference. 

Such an improvement using a relatively simple deterministic function led us to wonder whether we could go even further in improving performance. More precisely, considering Word translation as a retrieval task, the framework implemented was that of a ranking problem. In order to find the right translation, it was important to optimally rank potential candidates. This naturally led us to want to clearly define this ranking problem and to use the state of the art research on raking to tackle it. In this framework, the use of simple deterministic criteria such as NN, CSLS or ISF was a low-tech answer and left a large field of potential improvement to be explored. 

However, we wanted to keep the unsupervised framework, hence the idea of training the learning to rank algorithms on the learning of the translation of a language pair, English-Spanish for instance, assuming the existence of a dictionary. This would then allow us to apply the learning to rank algorithm for another language pair without dictionary, English-Italian for instance. Similarly to the case of CSLS, the criterion can be tested first at the end of the alignment carried out thanks to the Procustes-Wasserstein method. Then, in a second step, it can be integrated directly through the loss function in the alignment step. The following will quickly present the learning to rank framework in order to understand our implementation in more detail. 

\section{Learning to rank} \label{sec:ltr}

A ranking problem is defined as the task of ordering a set of items to maximize the utility of the entire set. Such a question is widely studied in several domains, such as Information Retrieval or Natural Language Processing. For example, on any e-commerce website, when given a query "iPhone black case" and the list of available products, the return list should be ordered by the probability of getting purchased. One can start understanding why a ranking problem is different than a classification or a regression task. While their goal is to predict a class or a value, the ranking task needs to order an entire list, such that the higher you are, the more relevant you should be.

\subsection{Theoretical framework}

Let's start by diving into the theoretical framework of Learning to Rank. Let $\psi := \{ (X,Y) \in \mathcal{X}^n \times \mathbb{R}^{n}_{+}\}$ be a training set, where: 

\begin{enumerate}
    \item $X \in \mathcal{X}^n$ is a vector, also defined as $(x_1,...,x_n)$ where $x_i$ is an item.
    \item $Y \in \mathbb{R}^{n}_{+}$ is a vector, also defined as $(y_1,...,y_n)$ where $y_i$ is a relevance labels.
    \item $\mathcal{X}$ is the space of all items.
\end{enumerate}

Furthermore, we define an item $x \in \mathcal{X}$ as a query-documents pair $(q,d)$.

The goal is to find a scoring function $f : \mathcal{X}^n \rightarrow \mathbb{R}^{n}_{+}$ that would minimizes the following loss : 

\begin{equation*}
    \mathcal{L}(f) := \frac{1}{|\psi|} \sum_{(X,Y) \in \psi} l(Y,f(X))
\end{equation*}

where $l : (\mathbb{R}^{n}_{+})^2 \rightarrow \mathbb{R}$ is a local loss function. 

One first and very important note is how $f$ is defined. This could be done in two ways: 

\begin{enumerate}
    \item We consider $f$ as a \textbf{univariate} scoring function, meaning that it can be decomposed into a per-item scoring function with $u : x \mapsto \mathbb{R}_+$. We will have $f(X) = [u(x_0), \cdots , u(x_n)]$. 
    \item We consider $f$ as a \textbf{multivariate} scoring function, meaning that each item is scored relatively to every other item in the set, with $f$ in $\mathbb{R}^{n}_{+}$. This means that changing one item could change the score of the rest of the set.
\end{enumerate}
While the first option is simpler to implement, the second one is much closer to reality, as an item's relevance often depends on the distribution its in. For example, an article's relevance to an e-commerce query will always depend on what the website offers you next to it.

We now have to define some metrics in order to judge how good a ranking is. We start by defining the \textbf{Discounted Cumulativ Gain} (DCG) of a list:

\begin{equation*}
    \text{DCG}@k(\pi,Y) := \sum_{j=1}^{k} \frac{2^{y_j} - 1}{\log_{2}(1+\pi(j))}
\end{equation*}

where: 
\begin{itemize}
    \item $Y = (y_1,...,y_n)$ are the ground truth labels
    \item $\pi(j)$ is the rank of the j-th item in f(X)
    \item $\frac{1}{ln_{2}(1+\pi(j))}$ is the discount factor
    \item $k$ is how much we want to go deep into the list. A low value of $k$ means that we want to focus on how well ranked the start of our list is.
\end{itemize}
Most of the time however we want to compare this metric to the DCG obtained from the ground truth labels. We then define: 

\begin{equation*}
    \text{NDCG}@k(\pi,Y) := \frac{DCG@k(\pi,Y)}{DCG@k(\pi^*,Y)}
\end{equation*}

where $\pi^*$ is the item permutations induced by Y.

\subsection{Loss functions}

In

\begin{equation*}
    \mathcal{L}(f) := \frac{1}{|\psi|} \sum_{(X,Y) \in \psi} l(Y,f(X))
\end{equation*}

we defined $l$ as a loss function between two ordered sets of items. One approach could be to use the metrics defined above, but as they are non-differentiable, this is not a feasible choice. Therefore, we have to develop some sort of surrogate loss function, differentiable, and with the same goal as our metric. Before diving into the possible approaches, one must define what pointwise, pairwise, and listwise loss functions are.

A pointwise loss will only compare one predicted label to the real one. Therefore, each item's label is not compared to any other piece of information. 
A pairwise loss will compare 2 scores from 2 items at the same time. With the pairwise loss, a model will minimize the number of pairs in the wrong order relative to the proper labels. 
A listwise loss can capture differences in scores throughout the entire list of items. While this is a more complex approach, it allows us to compare each score against the others. 

Let's give an example for each sort of loss function, starting with a pointwise one. The sigmoid cross entropy for binary relevance labels can be defined as:

\begin{equation*}
    l(Y,\hat{Y}) = \sum_{j=1}^{n} y_j\log(p_j) + (1-y_j)\log(1-p_j)
\end{equation*}

where $p_j = \frac{1}{1 + e^{-\hat y_j}}$ and $\hat Y$ is the predicted labels for each item from one model.

The pairwise logistic loss is a pairwise loss function that compares if pair of items are ordered in the right order. We can define it as: 

\begin{equation*}
    l(Y,\hat{Y}) = \sum_{j=1}^{n} \sum_{k=1}^{n} \mathbb{I}_{y_j > y_k} \log(1 + e^{\hat{y_k} - \hat{y_j}})
\end{equation*}

where $\mathbb{I}_{x}$ is the indicator function.

Finally, a listwise loss function like the Softmax cross-entropy can be defined as:

\begin{equation*}
    l(Y,\hat{Y}) = \sum_{j=1}^{n} y_j \log\left(\frac{e^{\hat{y_j}}}{\sum_{k=1}^{n} e^{\hat{y_k}}}\right)
\end{equation*}

All of these loss functions are surrogates that try to capture the goal of our metrics. Another approach would be to define a listwise loss function, as close as our metrics as possible. For example, one could try to build a differentiable version of the NDCG.

\subsubsection{ApproxNDCG: a differentiable NDCG}

Let's take a query $q$, and define several useful functions: $\delta: x \in \mathcal{X} \mapsto \delta(x) \in \{0,1\}$ the relevance of an item $x$ regarding of the fixed query $q$, $\pi: x \in \mathcal{X} \mapsto \pi(x) \in [1, \# \mathcal{X}]$ the position of $x$ in the ranked list $\pi$ and finally, $\mathbb{I}_{\{\pi(x) \leq k\}}$ still represents the indicator function of the set $\{\pi(x) \leq k\}$. The idea is to use an differentiable approximation of the indicator function. One can show that we have the following approximation:

\begin{equation*}
    \mathbb{I}_{\{\pi(x) \leq k\}} \approx \frac{e^{-\alpha (\pi(x) - k)}}{1 + e^{-\alpha (\pi(x) - k)}}
\end{equation*}

where $\alpha$ is a hyperparameter.

For a fixed query, we can re-define the DCG metric with the following equality:

\begin{equation*}
  \text{DCG}@k(\pi) := \sum_{x \in \mathcal{X}} \frac{2^{\delta(x)} - 1}{\log_{2}(1+\pi(x))} \mathbb{I}_{\{\pi(x) \leq k\}}
\end{equation*}

We now have to get an approximation of the $\pi$ function. The idea here is to get back to an indicator function since it is possible to compute them. We will be using the following equality:

\begin{equation*}
  \pi(x) = 1 + \sum_{y \in \mathcal{X}\backslash\{x\}} \mathbb{I}_{\{s_x \leq s_y\}}
\end{equation*}

where $s_x$ is defined as the score given to $x$ according to $f$, and define an approximation of $\pi$ with:

\begin{equation*}
  \hat{\pi}(x) = 1 + \sum_{y \in \mathcal{X}\backslash\{x\}} \frac{e^{-\alpha (s_x - s_y)}}{1 + e^{-\alpha (s_x - s_y)}}
\end{equation*}

We can now define our differentiable version of the DCG metric by using these approximations.

\section{RUBI: Ranked Unsupervised Bilingual Induction} \label{sec:semi_sup}

\textbf{Motivations:} Let's describe more precisely the functioning of our algorithm, denoted RUBI, although already mentioned in previous sections. Two points guided our approach:
\begin{enumerate}
    \item From a linguistic point of view, there is obviously a learning to learn phenomenon for languages. We observe that by assimilating the structure of the new language, its grammar and vocabulary to one of the already known languages, it is easier for us to create links that help learning.
    It is the search for these links that motivates us and we are convinced that they can be useful when inferring vocabulary. 
    \item improvement induced by the use of the CSLS criterion suggests that there are complex geometrical phenomena (going beyond the above-mentioned existence of hubs) within the representations of languages, both ante and post-alignment. Understanding these phenomena can lead to greatly increased efficiency. 
\end{enumerate}
\textbf{Framework:} Our goal is the same as for unsupervised bilingual alignment: we have a source language A and a target language B with no parallel data between the two. We want to derive an A-B dictionary, a classic BLI task. The specificity of our study is to assume that we also have a C language and an A-C dictionary at our disposal. To set up the learning to learn procedure, we proceed in 2 steps:
\begin{enumerate}
    \item \textbf{Learning:} Using the Procustes-Wasserstein algorithm, we align languages A and C in an unsupervised way. We then build a corpus of queries between the words from language A known from our dictionary and their potential translation into language C. Classical methods proposed the translation that maximized the NN or CSLS criteria. In our case, we use deep learning as part of our learning to rank framework to find a more complex criterion. One of the innovative features of our work is therefore to allow access to a much larger class of functions for the vocabulary induction stage. A sub-part of the dictionary is used for cross-validation. The way of rating the relevance of the potential translations, the inputs of the algorithm, the loss functions are all parameters that we studied and that are described in the next section. 
    \item \textbf{Prediction:} We thus have at the end of the training an algorithm taking as input a vocabulary word, in the form of an embedding as well as a list of potential translations. The output of our algorithm is the list sorted according to the learned criteria of these possible translations, the first word corresponding to the most probable translation and so on. We first perform the alignment of languages A and B using again the Procustes-Wasserstein algorithm. In a second step, thanks to the learning to rank, we perform the lexicon induction step. 
\end{enumerate}

\textbf{Choices and expected results:}A number of important points should be made:
\begin{enumerate}
    \item We assume that C's learning from A is of interest to B's learning. The truthfulness of this hypothesis is an interesting fact that we will be able to study. Realistically, it seems that learning Chinese from English will help us less to learn Italian than if we had chosen Spanish in the first place. However, it gives us a practical measure of proximity between languages, which is not obvious to infer at first glance and can be very interesting.
    \item We choose to use the Procustes-Wasserstein algorithm in the translation stage because we believe that maintaining the alignment method throughout our study allows us to truly evaluate its effectiveness and the specific geometric changes it induces. Since we assume that we have a dictionary at our disposal, we could use supervised methods, but they would not work in the same way.
    \item The multi-alignment scenario does not assume the existence of this dictionary. Realistically, however, it is consistent with existing use-case to assume that it is available. 
\end{enumerate}

Finally, a final conceptual point is important to raise. In the context of the CSLS criterion, we have seen in the above that its use after alignment has improved. However, actually incorporating it in the alignment phase by modifying the loss function has allowed for greater consistency and a second improvement. However, these two changes were separated. Yet, the learning to rank framework is quite different. The main reason is the non-linearity resulting from deep-learning, unlike CSLS. The global optimization is therefore much more complex and does not allow a relaxation to get back to a convex case. However, it is an area for improvement to be considered very seriously for future work.

\section{Experiments and Results} \label{sec:results}

\subsection{Unsupervised word translation}

\textbf{Implementation details:}
The general parameters used are described bellow. They are studied in more depth in the following subsection.
\begin{enumerate}
    \item Fasttext Word embeddings, learned on Wikipedia corpus, dimension 300, size 200 000 words. Dictionary of the 5,000 most frequent words for training and assessment, dictionary of the next 1,500 words for cross validation. 
    \item Alignment: Procustes-Wasserstein algorithm, 5 epoch of 5000 iterations. Learning rate of 0.5. Batch size of 500.
    \item Learning to Rank: 100000 iterations, batch size de 32, dropout rate of 0.5 for regularization, 3 hidden layers of size 256, 128 and 64, Adagrad Optimizer, group size of 4, queries of 10 potential translations (selected using the NN) for each 5000 words in dictionary. 11 Features (Similarity and CSLS(i) for i in [1,10] relative to the query word).
\end{enumerate}
\textbf{Baselines:}
We compare our method with Wasserstein-Procrustes (Wass. Proc.) \citet{bib2}, as well as two unsupervised approaches: the adversarial training (adversarial) of Conneau et al.\citet{bib5} and the Relaxed CSLS loss (RCSLS) \citet{bib27}. All the numbers are taken from their papers.
The reference benchmark is the translation from English to French, Spanish, Russian and German as well as the reverse translation. 
\newline
\textbf{Mains results:}
In order to quantitatively assess the quality of each approach, we consider the problem of bilingual lexicon induction. Following standard practice, we report the precision at one. Contrary to other methods, we use an auxiliary language, which we show in brackets next to the result. We then specifically study this choice in relation to the translation. The general parameters used are described above. Table 1 summarizes our results, compared to existing state of the art on reference BLI task. Our method outperforms all existing methodology, with a vast margin. Hence, Our assumption that learning a language for training purposes brings a lot is confirmed. Without even going into the alignment phase, our criterion brings a real gain for unsupervised translation, for a reasonable computational complexity (about fifteen minutes per translation).
the major contribution is for English-Russian translation, with a gain of 35\% compared to the best existing method (Russian-English translation is not included because we had no access to dictionaries required for the learning phase). 

\begin{table}
\begin{center}
\begin{tabular}{lcccccccc}

\toprule

            Method & EN-ES & ES-EN & EN-FR & FR-EN \\

\hline 

Wass. Proc. - NN   &  77.2  & 75.6 & 75.0 & 72.1  \\

Wass. Proc. - CSLS & 79.8 & 81.8 & 79.8 & 78.0  \\

Wass. Proc. - ISF  & 80.2 & 80.3 & 79.6 &  77.2   \\

\hline 

Adv. - NN          & 69.8 & 71.3 & 70.4 & 61.9  \\

Adv. -CSLS         & 75.7 & 79.7 & 77.8 & 71.2  \\

\hline

RCSLS+spectral     & 83.5 & 85.7 & 82.3 & 84.1  \\

RCSLS              & 84.1 & 86.3 & 83.3 & 84.1  \\

\hline 

RUBI               & \textbf{93.3} (\textit{DE}) & \textbf{91.6} (\textit{FR}) & \textbf{93.8} (\textit{NL}) & \textbf{91.9} (\textit{IT})  \\

\hline
\hline
\hline 

             & EN-DE & DE-EN & EN-RU & RU-EN \\

\hline 

Wass. Proc. - NN   & 66.0 & 62.9 & 32.6 & 48.6  \\

Wass. Proc. - CSLS & 69.4 & 66.4 & 37.5 & 50.3  \\

Wass. Proc. - ISF  & 66.9 & 64.2 & 36.9 & 50.3  \\

\hline 

Adv. - NN          & 63.1 & 59.6 & 29.1 & 41.5  \\

Adv. -CSLS         & 70.1 & 66.4 & 37.2 & 48.1  \\

\hline

RCSLS+spectral     & 78.2 & 75.8 & 56.1 & 66.5  \\

RCSLS              & 79.1 & 76.3 & 57.9 & 67.2  \\

\hline 

RUBI               & \textbf{93.6} (\textit{HU}) & \textbf{89.8} (\textit{FR}) & \textbf{83.7} (\textit{HU}) & -  \\

\bottomrule

\end{tabular}
\caption{Benchmark Results for Bilingual Lexicon Induction}
\end{center}
\end{table}

Table 2 summarizes for French, Italian, Russian, Portuguese and German the BLI for translation to English. These were the only languages for which we had dictionaries between each pair. This allowed us to set up a learning step without English being one of the two languages, which was impossible for the more complete list of languages we used afterwards. The best result is the one shown above in Table 1.

\begin{table}
\begin{center}
\begin{tabular}{lccccc}

\toprule

            Pivot & ES-EN & FR-EN & DE-EN & PT-EN\\

\hline 

ES   & - & 91.6 & 83.3 & 88.8  \\

FR   & \textbf{91.6} & - & \textbf{89.8} & \textbf{89.8} \\

IT   & 91.3 & \textbf{91.9} & 88.5& 89.4\\

DE   & 90.0 & 91.5 &  - & 89.8 \\

PT   & 91.4 & 91.6 & 89.8 & -   \\

\bottomrule
\end{tabular}
\caption{BLI efficiency for reverse translation to English}
\end{center}
\label{fig:rev_main}
\end{table}

\subsection{Ablation Study}

In this section, we evaluate the impact of some of our implementation choices on the performance of RUBI. We focus in particular on the loss function, the query size, the group size as well as the features and relevance system used . 

\subsubsection{\textbf{Impact of Loss function:}}
As described above, the learning to rank framework allows the use of numerous loss functions corresponding to different scenarios (0-1 relevance, pointwise, listwise and pairwise...). In our case, we wanted to maximize the BLI criterion, i.e. to have the maximum performance just for the top of the list in order to be able to compare our work with the existing literature. Other objectives can be considered, such as maximizing the presence of the right translation among the first X suggestions and then manually look for the correct translation in this subset. We have therefore added the BLI criterion among the other criteria (NDCG, RP, RR...). The closest existing one was NDCG@1 (1st position). We then tested a large majority of the existing loss functions to see which one was the most efficient given our objective. The graph below presents the results for 5 of these functions, representative of the different existing categories. The evaluation criterion is the BLI for EN-ES translation. The two functions that perform best are the Approximate NDCG loss (which maximizes a differentiable approximation of NDCG) and the MLE loss list (which maximizes the likelihood loss of the probability distribution). In other experiments, we found that these two loss functions continue to perform similarly, hence our default choice of Approx NDCG. We have also shown this function with a group size of 1 and 2 (parameter described in the learning to rank section).

\begin{figure}[ht]
\begin{center}
\includegraphics[scale=0.4]{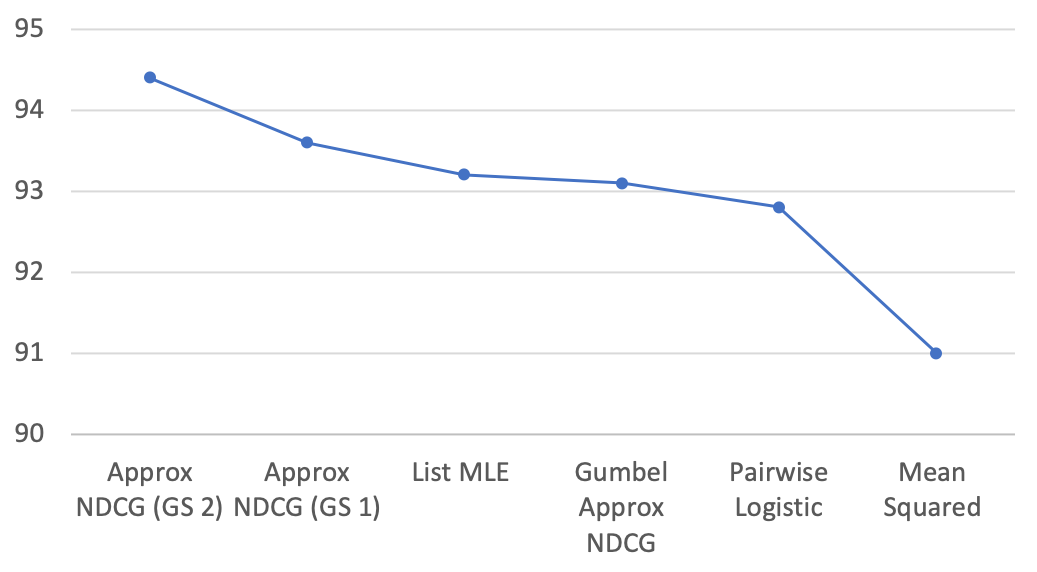}
\end{center}
\caption{Loss function Impact.} 
\end{figure}

\subsubsection{\textbf{Impact of group size:}}
As seen above, group size has a great influence on our criterion. This is quite consistent given our concern for ranking compared to other embeddings. The dilemma is however to optimize the computation time because increasing the group size exponentially increases the number of calculations. The graph below presents the BLI criterion for EN-ES, using two loss functions and varying the group size. A clear improvement can be observed. However, the biggest increase seems to occur when changing the group size from 1 to 2 and then the curve stabilizes. Therefore, in our computations, we used mainly a group size of 2 and often of 4 when looking for greater accuracy. 
\begin{figure}[ht]
\begin{center}
\includegraphics[scale=0.4]{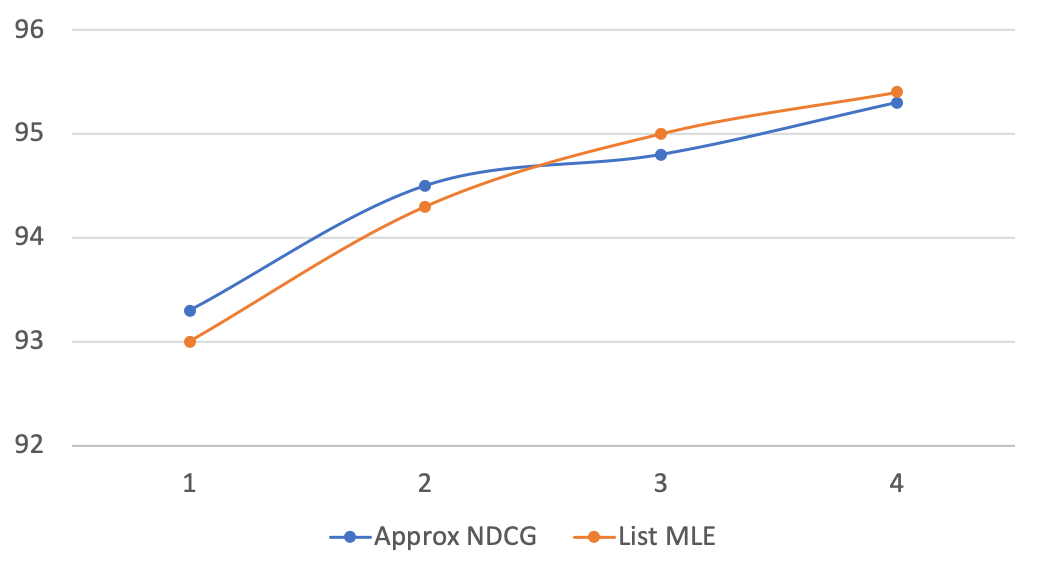}
\end{center}
\caption{Group Size Impact.}
\label{fig:gogo1}
\end{figure}

\subsubsection{\textbf{Impact of feature system:}}
As stated above, our framework receive as input a query list i.e. for each word in the dictionary, a list of potential translation. For each of these potential translation, we compute a relevance label (estimated using ground truth) and a list of features. The relevance label is used only for the training and we be studied bellow. The features for the each potential translation in a query can incorporate several elements:
\begin{enumerate}
    \item the word embedding of the potential translation (size 300)
    \item the word embedding of the query (size 300)
    \item pre-computed features such a distance to query word in the aligned vector space, CSLS distance, ISF....
\end{enumerate}
Those features are crucial for the learning as it will fully rely on it. 
At first, we decided to only use the word embedding of the potential translation and of the query. That gave us a 600 feature list. However, after several experiments, we noticed that the learning to rank algorithm, despite the variation of the parameters, was not able to learn relevant information from these 600 features, the performance was poor. The function learned through deep learning was less efficient than a simple Euclidean distance between the potential translation and the query (NN criterion). In fact, after consulting the literature, we realised that using such a number of features is not very common. Most algorithms were only using pre-computed features (often less than a hundred). Although this information is already interesting in itself, we therefore turned to the second approach. We chose to restrict ourselves to certain well-specified types of pre-computed features in order to evaluate their full impact. More precisely, for a fixed k parameter, we provided as features the euclidean distance to the query, as well as the CSLS(i) "distance" for i ranging from 1 to \(k\). In other words, we provided information about the neighborhood through the penalties described in the section on CSLS. In the context of this work, we did not want to use other features (ISF in particular) to focus specifically on the contribution of CSLS but this is an easy improvement path to exploit for future work.
As outlined above, this simple framework allows a considerable improvement of the BLI. Below, we describe the evolution of the BLI for EN-ES translation by varying the \(k\) parameter. \(k\)=0 correspond to the use only of euclidean distance to the query. We observe a major increase for \(k\) from 0 to 1, a lesser increase for \(k\) from 1 to 4-5 and stabilization thereafter, with a slight maximum towards \(k\)=14. 

This leads us to believe that the relevant information for the algorithm is just in the close neighborhood of the point (i.e. the few closest neighbors) and the addition of features describing a more distant neighborhood brings only marginal information. This preliminary conclusion is clearly worth further study. Indeed, the functioning of the neural network at the very heart of the learning to rank algorithm is that of a black box and this conclusion deserves to be reinforced. Its veracity would allow a better understanding of the phenomenon of hubbness in the context of word embeddings clouds. 
\begin{figure}[ht]
\begin{center}
\includegraphics[scale=0.4]{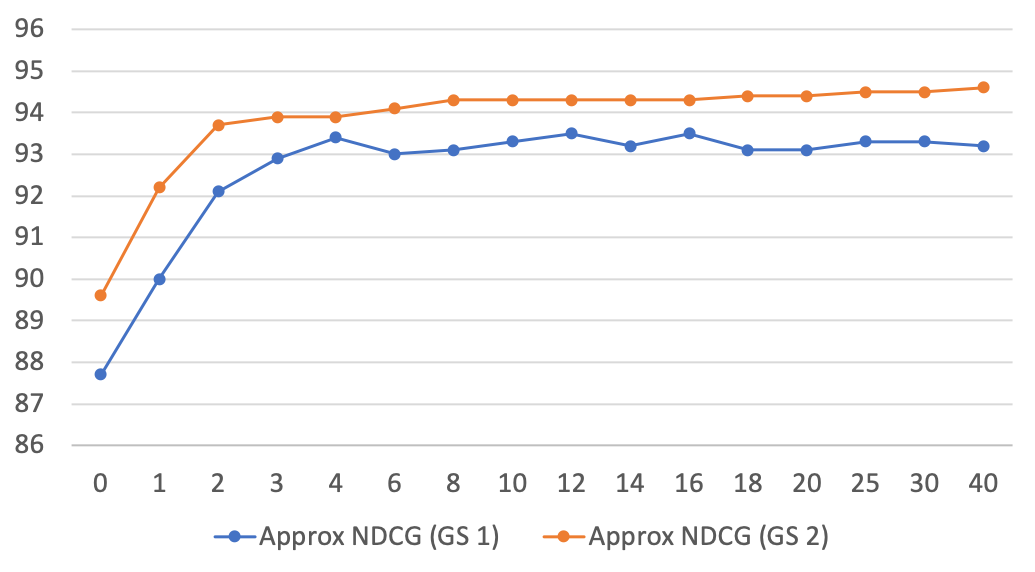}
\end{center}
\caption{CSLS feature Impact.}
\label{fig:gogo2}
\end{figure}

\subsubsection{\textbf{Impact of query relevance:}}
The learning to rank framework used makes it possible to rate the relevance of a word embedding in relation to a query thanks to a system of integers. A relevance of 0 translates a non relevance, then the higher the relevance is, the more the word embedding will be relevant i.e. close to the translation. For the ranking of the list of potential translations, it is these relevancies that will be used. For the pointwise loss function (i.e. regression problem), it will be a matter of predicting these labels. In the peerwise and listwise functions, it is more complex. We have considered 3 relevance scenarios. This relevance setting is the key of a good algorithm as it represent the function we aim to learn. With this in mind, we tried three different options:
\begin{enumerate}
    \item \textbf{Binary Relevance:} The correct translation and its synonyms have a relevance of 1, the other candidates have a relevance of 0 (not relevant). This system allows to use some specific loss functions very efficiently in these scenarios (like sigmoid cross entropy loss). However, the context of our study was not very appropriate for this relevance system. Indeed, we often had only one correct translation per query, for about ten erroneous translations. The algorithm was therefore not able to learn about the corpus of good translations and almost systematically predicted that the translation was erroneous, i.e. a relevance of 0. 
    \item \textbf{Continuous Relevance:} On the other hand, we asked ourselves how to translate the relevance of a potential candidate for translation, even if it is not the right translation or a synonym. In other words, how do you extract the information it contains? The previous approach only considered the question: "is this a correct translation ?" instead of asking "to what extend is this a correct translation ?" In this context, we had 3 suggestions:
    \begin{enumerate}
        \item \textbf{Intra-distance:} The core idea of word embeddings is to translate the contextual proximity of 2 words into a proximity in terms of distance in a space. Therefore, if we are interested in the relevance of a word ("dog") in the context of a translation ("chat"), we can consider the distance between the embeddings of the candidate ("dog") and the correct translation ("cat"). We talk about intra-distance because it's a distance in the target space.  When these distances are calculated, we can then classify the words by proximity and give them a label of relevance thanks to this. Although appealing on paper, this approach led to poor results for a hidden reason: the learning to rank algorithm sought to maximize the ordering of the response list. When it was given too many distinct possible labels, it made it easier to make a large number of correct predictions about the relative ranking in the answer list. Maximizing the top of the list, which is our main interest, was no longer a priority at all. The other concern was that in very large dimensions, the notion of distance also lost its meaning. It was therefore less relevant to note these distances.
        \item \textbf{Extra-distance:} In a logic close to the previous method, we wondered if it could be interesting to consider the distances in the source space instead of those in the target space. In other words, instead of looking at the distance between our candidate ("dog") and the correct translation ("cat") of the target word ("chat"), we could look at the translation of the candidate in the source space ("chien") and look at the distance between this translation and the word to be translated. This idea was therefore based on the word embeddings in the source space and did not use the word embeddings in the target space. Unfortunately, we didn't have a dictionary for each of the 200,000 words used, so this passage in the source space was complicated to settle. 
        \item \textbf{Exogenous distance:} We finally wondered if we could incorporate an exogenous distance thanks to external algorithms that could provide a proximity between words other than the one given by the embeddings (syntactic, etymological...). This track remains in our suggestions for improvement but the poor results obtained with the intra-distance method dissuaded us from exploring it right away. 
    \end{enumerate}
    \item \textbf{Semi-binary:} The solution chosen for the implementation was a trade off between the two previous approaches: the correct translation and its synonyms receive a label of 2 and the other words receive a label of 1 (very low relevance). This pushes the algorithm to focus as much as possible on the correct translation while using the above information in the words (to a lesser extent).
\end{enumerate}

\subsubsection{\textbf{Impact of query size:}}
Finally, we studied the impact of the size of the query i.e. the number of potential translations provided for each word. A large number of potential translations gives you more choice but the risk of getting it wrong is greater. One must also ask whether the algorithm is able to learn in a relevant way if it is provided with a large amount of information (given that a semi-binary relevance system is used). The experience setting is the same as for the previous points (BLI induction for EN-ES). There is a low incidence of the number of queries on the results, a very slight but perceptible decrease. The algorithm is therefore able, despite a large number of candidates, to discern the correct information. 
\begin{figure}[ht]
\begin{center}
\includegraphics[scale=0.4]{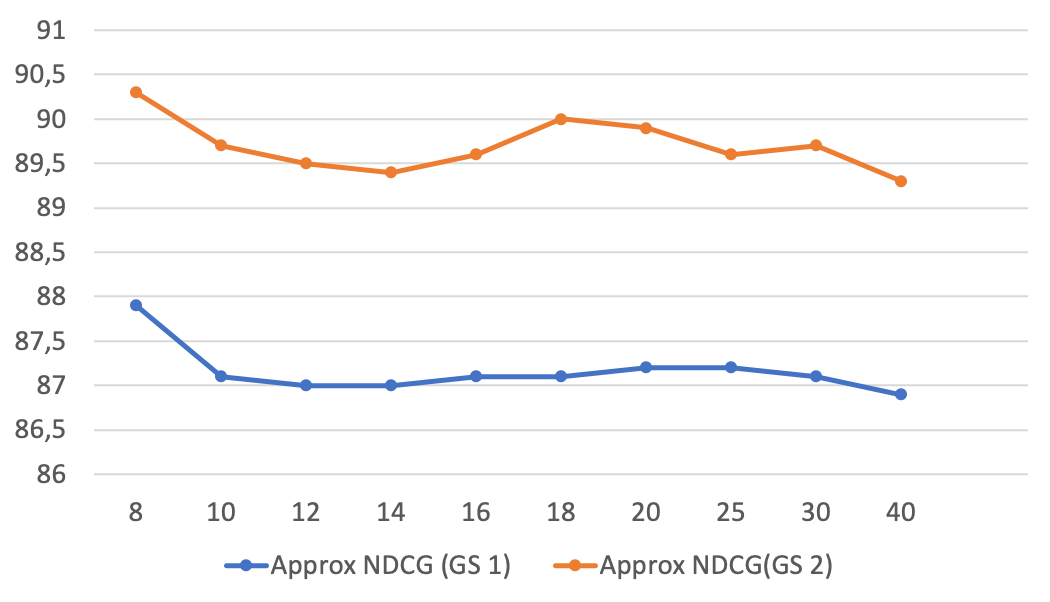}
\end{center}
\caption{Query Size Impact.}
\label{fig:gogo3}
\end{figure}

\subsection{Impact of learned idiom in inference}

\textbf{Experiment details:} In this section, we evaluate the impact of the idiom used for learning when it comes to translation. As it is very costly to compute, we only used our 4 reference languages as target to begin: Spanish (ES), French (FR), German (DE) and Russian (RU). More computation are currently on their way. Facebook data gave us access to dictionary from English to more than 20 idioms. The following table present the BLI criterion for the translation using the idiom in the left for the learning process. The last line is the benchmark for the translation associated with this pair, using one of the other previously described unsupervised learning methodology. 
The last columns deals with the learning part: the first digit is the is the BLI criterion obtain for the learning language at the end of the training. The second one is the BLI criterion for the learning language obtained when aligning and translating this idiom with English, using Wasserstein Procrustes Unsupervised Alignment and CSLS criterion)\citet{bib2}. If the first number does not appear, it means that the alignment is of poor quality and that the learning step leads to blatant over-feating, so the number makes little sense. 

\textbf{Mains results:} Several conclusions can be drawn from this table:
\begin{enumerate}
    \item \textbf{Prediction stability:} There is a strong stability for the prediction of a given language. Except for languages very poorly aligned with English, the results seem to be globally identical when the idiom used for learning varies (difference of less than 1\%). This leads to more remarks. Although the learning to rank algorithm is a black box, it seems that it always leads to a similar criterion at the end of the learning process, a criterion much more powerful than those used so far (NN, CSLS, ISF among others). The proximity between the learning language and the prediction is not in play, it is a very strong result. The findings in the next section support this observation. This reinforces the hypothesis of the similarity of the word cloud structure and thus legitimizes the global approach of the alignment implemented. The performance obtained during the learning process does not seem to be correlated with the performance of the predictions either. We can also ask ourselves whether using several languages during the learning phase is really worthwhile. This seems to be a logical next step for the project, but this remark suggests a lesser increase in terms of efficiency. 
    \item \textbf{Impact of Target Language:} There is, however, a strong variation in the BLI criterion depending on the language to be predicted. This point will be studied in more detail later. Yet, this variation seems to be positively correlated with the quality of the alignment of this language with English. Thus, Russian performs less well than German or French in this process. However, these results should be highlighted: the greatest contribution of our method is precisely for Russian when compared to existing benchmarks. We observe a gain of more than 25\%, while those for French, Spanish and German are around 10\%.
    \item \textbf{Learning step:} The training performance (last columns) seems to depend directly on the quality of the alignment of the language used for learning with English. Figure 8 plots the BLI criterion in the training step according to the CSLS criterion, i.e. the quality of the alignment of the language used for learning with English. The trend that emerges is that of a very clear positive correlation (linear trend plotted in red, \(\mathbf{R}^2= 0.82\)). We have also shown the averages per language family (Romance, Germanic and Uralic). In conclusion, it seems easier to learn using a language that is well aligned with English. Although this seems logical, it is not that obvious. Three clusters seem to appear in conjunction with the different families. Romance languages are associated with a high rate of alignment with English and therefore with high performance in the learning stage. 
    The Germanic language cluster has a lower performance combined with a slightly lower quality alignment. Knowing that English belongs to the Germanic language type, it is interesting to note this slight underperformance in alignment compared to Romance. Finally, the Slave cluster shows the worst performance in terms of alignment with English and therefore also the worst for the learning step.   
\end{enumerate}

\begin{table}[h]
\begin{tabular}{lccccccccc}

\toprule

        Pivot idiom & EN-ES & EN-FR & EN-DE & EN-RU & \textit{Training}\\

\hline 
\hline 

\textit{Romance}            &  &  &  &  &  \\

\hline 

French              & 95.0 & - & 92.9 & 81.0 & \textit{92.5 / 80.2} \\

Italian             & 95.2 & 93.6 & 93.0 & 82.1 & \textit{89.6 / 76.3} \\

Portuguese          & 94.8 & 93.6 & 92.6 & 81.1 & \textit{91.3 / 81.3} \\

Spanish             & - & 93.7 & 93.1 & 81.5 & \textit{92.4 / 82.1} \\

Catalan             & 94.8 & 93.3 & 92.4 & 81.6 & \textit{87.0 / 63.4} \\

Romanian            & 94.9 & 93.5 & 92.1 & 80.7 & \textit{84.4 / 60.0} \\

\hline 

\textit{Germanic}            &  &  &  &  &  \\

\hline 

Dutch             & \textbf{95.3} & \textbf{93.8} & 93.0 & 81.4 & \textit{88.1 / 74.3} \\

German            & 94.6 & 93.0	& - & 82.7 & \textit{90.7 / 70.9} \\

Norwegian         & 95.2 & 93.7 & 93.0 & 81.1 & \textit{84.7 / 62.4} \\

Danish            & 95.0 & 93.4 & 93.1 & 81.6 & \textit{85.6 / 63.7} \\

Swedish           & 48.2 & 40.2 & 67.9 & 64.4 &  \textit{- / 0.2} \\

\hline 

\textit{Slavic}            &  &  &  &  &  \\

\hline 

Russian           & 95.2 & 93.6 & 93.5 & - & \textit{76.0 / 42.6} \\

Ukrainian         & 94.9 & 93.2 & 93.1 & 83.2 & \textit{77.4 / 32.4} \\

Slovak            & 94.2 & 92.5 & 92.8 & 81.7 &  \textit{- / 14.8} \\

Polish            & 95.1 & 93.4 & 93.5 & 82.6 & \textit{79.1 / 49.1} \\

Bulgarian         & \textbf{95.3} & 93.5 & 92.8 & 82.2 & \textit{80.3 / 47.1} \\

Czech             & 95.2 & 93.7 & 93.5 & 82.9 & \textit{78.9 / 49.7} \\

Croatian          & 94.9 & 93.5 & 93.1 & 82.7 & \textit{76.6 / 34.1 }\\

Slovenian         & 94.9 & 93.4 & 93.1 & 82.2 & \textit{79.1 / 33.9} \\

Macedonian        & 47.4 & 40.6 & 68.0 & 63.4 &  \textit{- / 0.6} \\

\hline 

\textit{Others}            &  &  &  &  &  \\

\hline 

Hungarian          & 95.2 & 93.7 & \textbf{93.6} & \textbf{83.7} & \textit{79.6 / 48.0} \\

Estonian           & 47.6 & 41.8 & 68.7 & 65.2 &  \textit{- / 0.4} \\

Greek              & 95.2 & 93.5 & 93.0	& 82.5 & \textit{83.5 / 48.5} \\

Arabic             & 94.5 & 93.4 & 93.3 & 82.6 & \textit{84.2 / 38.4} \\

Hebrew             & 94.7 & 93.4 & 93.1 & 82.3 & \textit{78.3 / 42.1} \\

Indonesian         & 94.7 & 93.1 & 92.7 & 80.6 & \textit{85.4 / 71.7} \\

Turkish            & 46.5 & 38.0 & 68.2 & 62.9 & \textit{- / 0.3} \\

Vietnamese         & 41.5 & 36.6 & 66.3 & 59.2 &  \textit{- / 0.1} \\

\hline 

\textit{Benchmark}             &  \textit{84.1}  &  \textit{83.3} &  \textit{79.1} &  \textit{57.9} &  \\

\bottomrule
\end{tabular}
\caption{BLI Results for English Translation using all available idioms }
\label{fig:gogo4}
\end{table}

\begin{figure}[h]
\begin{center}
\includegraphics[scale=0.4]{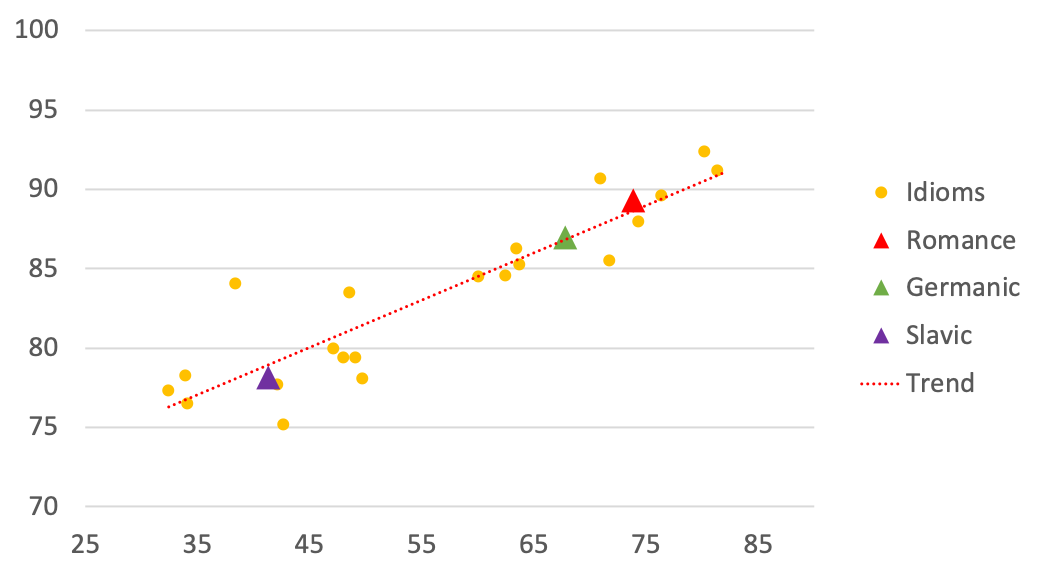}
\end{center}
\caption{Impact of Alignment Quality on Training.}
\label{fig:gogo5}
\end{figure}

\subsection{Impact of target idiom in inference}

\textbf{Experiment details:} Finally, we evaluate the impact of the idiom used for prediction when it comes to translation. We used 4 different languages for the learning step: French, Spanish, German and Russian. The previous section had shown that the language used for learning had little impact on prediction, hence the limited choice of languages. We then try to deduce for more than twenty languages the dictionary with English. Table 4 presents the results of this study. The languages to be predicted are in the left-hand column and those used for learning are in the first row. The last line is the benchmark used. This is again the BLI at the end of the alignment and reflects the quality of the alignment with English. 
\newline
\textbf{Mains results:} We can make the following observations:
\begin{enumerate}
    \item \textbf{Performances:} The algorithm is very efficient and allows to obtain a BLI criterion of at least 80\% up to 95\%, even more for all languages whereas the previous techniques were much less efficient. RUBI makes it possible to establish a new benchmark of quality to be surpassed in this task, which we hope can become a reference. The minimum gain observed is 10\%, which in the literature on the subject is considerable. 
    \item \textbf{Efficiency Forecasting:} We wondered whether we could quantify the contribution of our method in relation to the existing benchmark. Figure 9 shows the gains in terms of BLI points (e.g. +20\% corresponds to a change from a 60\% BLI benchmark to an 80\% BLI benchmark) as a function of the quality of the initial alignment, i.e. the benchmark. We observe a very strong negative correlation (linear trend plotted in red, \(\mathbf{R}^2= 0.96\)). This negative correlation is quite logical: it is easier to have a big BLI gain for languages initially misaligned with English. This line gives access to a first prediction, given the alignment of a language with English on the contribution that our method can give. This prediction seems very stable and relevant. Here again, we observe three clusters: the Romance and Germanic languages have initially a good alignment with English and thus present a relatively weak grain. The Slavic cluster gathers languages that are less well aligned with English and therefore has a lot to gain from our method. It also includes most of the category other languages.
    \item \textbf{Distant idioms:}It should be noted, however, that we did not use some accessible languages with poor alignment with English: Estonian (0.46\%), Macedonian (0.58\%), Swedish (0.24\%), Turkish (0.28\%) or even Vietnamese (0.08\%). These languages are interesting because they may represent the type of languages for which there is no parallel data with English. However, this is another type of scenario, which deserves to be studied separately. Some parameters need to be adapted for this distinct use case. For example, it is rarer that the correct translation is so close in the aligned space to the query word. It is necessary to look more for each query in the hundred candidates than in the ten, as we do at present. However, this is an exciting avenue to explore in the future. 
\end{enumerate}

\begin{table}[h]
\begin{tabular}{lccccccccc}

\toprule

        Learning idiom & Spanish & French & German & Russian & \textit{Bench.}\\

\hline 
\hline 

\textit{Romance}            &  &  &  &  &  \\

\hline 

French              & 93.7 &  -   & 93.0 & 93.6 & \textit{80.2} \\

Italian             & 91.5 & 91.8 & 91.4 & 91.6 & \textit{76.3} \\

Portuguese          & 94.0 & 93.8 & 93.8 & 93.9 & \textit{81.3} \\

Spanish             &  -   & 95.0 & 94.6 & 95.2 & \textit{82.1} \\

Catalan             & 87.3 & 87.1 & 85.3 & 86.8 & \textit{63.4} \\

Romanian            & 88.4 & 88.5 & 87.7 & 87.9 & \textit{60.0} \\

\hline 

\textit{Germanic}            &  &  &  &  &  \\

\hline 

Dutch             & 90.5 & 90.3 & 90.1 & 90.3 & \textit{74.3} \\

German            & 93.5 & 92.9	& -    & 93.5 & \textit{70.9} \\

Norwegian         & 86.1 & 85.4 & 86.0 & 86.6 & \textit{62.4} \\

Danish            & 89.4 & 89.7 & 89.0 & 89.8 & \textit{63.7} \\

\hline 

\textit{Slavic}            &  &  &  &  &  \\

\hline 

Russian           & 81.5 & 81.0 & 82.7 &  -   & \textit{42.6} \\

Ukrainian         & 82.3 & 80.2 & 82.7 & 83.3 & \textit{32.4} \\

Polish            & 86.2 & 85.6 & 86.3 & 87.3 & \textit{49.1} \\

Bulgarian         & 84.5 & 84.3 & 84.5 & 85.1 & \textit{47.1} \\

Czech             & 82.5 & 82.9 & 83.6 & 84.4 & \textit{49.7} \\

Croatian          & 78.4 & 77.8 & 77.9 & 78.8 & \textit{34.1 }\\

Slovenian         & 80.6 & 79.8 & 80.5 & 82.2 & \textit{33.9} \\

\hline 

\textit{Others}            &  &  &  &  &  \\

\hline 

Hungarian          & 79.4 & 79.5 & 79.6 & 79.6 & \textit{48.0} \\

Greek              & 86.0 & 86.0 & 86.6 & 87.3 & \textit{48.5} \\

Arabic             & 87.1 & 85.4 & 87.4 & 88.6 & \textit{38.4} \\

Hebrew             & 79.4 & 79.0 & 79.6 & 80.3 & \textit{42.1} \\

Indonesian         & 90.2 & 89.9 & 90.1 & 89.9 & \textit{71.7} \\

\bottomrule
\end{tabular}
\caption{BLI Results for all idioms Translation using ES, FR, DE and RU for learning}
\label{fig:gogo4}
\end{table}

\begin{figure}[ht]
\begin{center}
\includegraphics[scale=0.4]{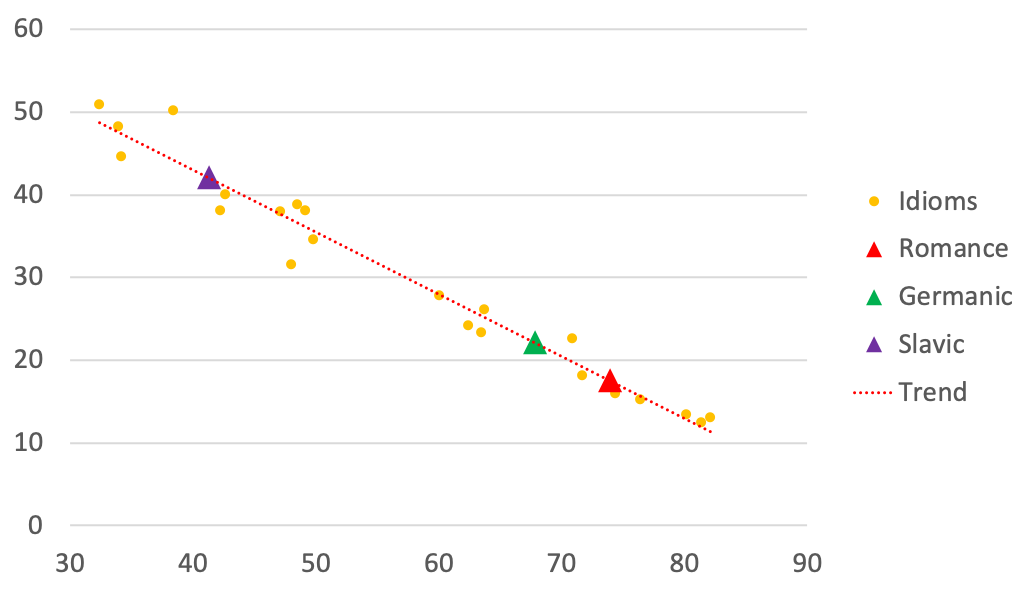}
\end{center}
\caption{Gain (points of BLI) after the use of RUBI in function of the quality of alignment.}
\label{fig:gogo6}
\end{figure}

\section{Conclusion}

This paper formulate a new approach to the unsupervised bilingual lexicon induction problem, using learning to rank tools. Our approach, leveraging knowledge of previous idioms acquisition, significantly improves the quality of the induction, outperforming state of the art methodology and setting a new benchmark. As a result, it produces high-quality dictionaries between different pairs of languages, with up to 93.8\% on the Spanish-French word translation task.  Moreover, we point out the stability of the prediction when the idiom used for learning varies, a strong argument in favor of the similarity of the structure of word embeddings clouds. 

\bibliography{references}
\bibliographystyle{jfm}
%
\appendix
\onecolumn
\section{Online Software Resources} \label{sec:software}

To make the discussed results useful and reproducible, our code and the software resources used are freely available online.\\

\begin{enumerate}

	\item The code, main examples and the files used for tables and graphs are accessible at \url{https://github.com/Gguinet/semisupervised-alignement.git}\\
	      
	\item It is still a private directory, however, and we are currently reformatting the code so that it can be made publicly available in the coming days. All you have to do is ask us for permission to add you to it if you can't access it. \\

	\item All data used (word embeddings and dictionary) are coming from Facebook public files on the topic. A part of it can be found using the link \url{https://fasttext.cc}.\\

	\item As the simulations were very demanding in term of computed power, we used Google Compute Engine, with the following settings: 8 virtual processors, n1-highmem-8, high memory capacity, 500 Go of memory, Ubuntu, Version 18.04. More information on how to use it is on the github file of the project.\\

\end{enumerate}

\end{document}